\documentclass[conference]{IEEEtran}
\IEEEoverridecommandlockouts
\usepackage{cite}
\usepackage{amsmath,amssymb,amsfonts}
\usepackage{algorithmic}
\usepackage{graphicx}
\usepackage{textcomp}
\usepackage{xcolor}
\usepackage{multirow}
\usepackage{multicol}
\usepackage{svg}

\usepackage{graphicx}
\usepackage{tikz}
\tikzset{mtm/.style={
  remember picture,
  inner sep=0pt,
  outer ysep=0.1em,
}}

\def\BibTeX{{\rm B\kern-.05em{\sc i\kern-.025em b}\kern-.08em
    T\kern-.1667em\lower.7ex\hbox{E}\kern-.125emX}}
\begin{document}

\title{Enabling Quantum Natural Language Processing for Hindi Language \\}

\author{
        \IEEEauthorblockN{Naman Srivastava}
	\IEEEauthorblockA{\textit{Department of DSIS} \\
		\textit{IIIT Dharwad}\\
		India \\
		srinaman2@gmail.com} 
        \and
        \IEEEauthorblockN{Gaurang Subhash Belekar}
	\IEEEauthorblockA{\textit{Department of ECE} \\
		\textit{IIIT Dharwad}\\
		India \\
		belekargaurang@gmail.com}
	\and
	
	\IEEEauthorblockN{Gautami Sanjay Naik}
	\IEEEauthorblockA{\textit{Ph.D in Quantum Computing} \\
		\textit{IIIT Dharwad}\\
		India \\
		gautaminaik2000@gmail.com} \\
        \and
	\IEEEauthorblockN{Sunil Saumya}
	\IEEEauthorblockA{\textit{Department of DSIS} \\
		\textit{IIIT Dharwad}\\
		India \\
		sunil.saumya@iiitdwd.ac.in}
         \and
	\IEEEauthorblockN{\hspace{6.45cm}Aswath Babu H}
	\IEEEauthorblockA{\hspace{6.45cm}\textit{Faculty of Humanities and Sciences } \\
		\hspace{6.45cm}\textit{IIIT Dharwad}\\
		\hspace{6.45cm}India \\
		\hspace{6.45cm}aswath@iiitdwd.ac.in}
}

\maketitle

\begin{abstract}
Quantum Natural Language Processing (QNLP) is taking huge leaps in solving the shortcomings of classical Natural Language Processing (NLP) techniques and moving towards a more "Explainable" NLP system. The current literature around QNLP focuses primarily on implementing QNLP techniques in sentences in the English language. In this paper, we propose to enable the QNLP approach to HINDI, which is the third most spoken language in South Asia. We present the process of building the parameterized quantum circuits required to undertake QNLP on Hindi sentences. We use the pregroup representation of Hindi and the DisCoCat framework to draw sentence diagrams. Later, we translate these diagrams to Parameterised Quantum Circuits based on Instantaneous Quantum Polynomial (IQP) style ansatz. Using these parameterized quantum circuits allows one to train grammar and topic-aware sentence classifiers for the Hindi Language. 

\end{abstract}


\section{Introduction}

In the ever-evolving landscape of information technology and computational linguistics, Quantum Natural Language Processing (QNLP) represents a groundbreaking paradigm shift at the intersection of Quantum Computing and natural language understanding. As the world delves deeper into the realms of Quantum Mechanics, harnessing the power of Quantum Computing to enhance the processing and comprehension of human language has emerged as a promising frontier. QNLP is a burgeoning field that promises to revolutionize how we interact with and extract knowledge from textual data, opening up new horizons for applications ranging from machine translation and sentiment analysis to information retrieval and cognitive computing. The first high-level Python library for QNLP was presented in 2021 as an open-source toolkit \cite{kartsaklis2021lambeq}.

Traditional Natural Language Processing (NLP) has made remarkable strides in automating tasks like text classification, machine translation, and sentiment analysis. However, it still grapples with inherent limitations, such as the inability to process and comprehend the vast nuances of human language efficiently. On the other hand, the model that interprets the Quantum Mechanical Phenomena is found to be equivalent to the model of natural language, which entails QNLP to be a quantum native of language processing. The processing of linguistic structure can be easily encoded through a quantum regime, whereas encoding grammar in classical means is very tedious and costly.  

Quantum Computing, on the other hand, leverages the principles of superposition and entanglement, enabling it to handle complex computational tasks exponentially faster than classical computers. Performing Natural Language Processing (NLP) on Hindi text is significant due to the language's vast user base, cultural relevance, and economic opportunities. It can offer insights into sentiments, opinions, and societal nuances, aids in governance and policy analysis, supports content creation and localization, enhances education, and has applications in healthcare, media analysis, linguistic research, and disaster response.

In particular, the QNLP pipeline relies on diagrammatic representation, wherein it involves first parsing an input sentence followed by generating its associated string diagram. This diagram serves as the basis for constructing a trainable Quantum Circuit or a Parameterized Quantum Circuit (PQC). The same method is implemented here for the case of Hindi sentences. This research endeavour delves into the methodology for crafting a Parameterized Quantum Circuit from Hindi sentences, establishing a fundamental framework for the development and training of a Quantum Natural Language Model tailored to the Hindi language. This work is inspired by the recent work on pregroup representation of Language \cite{debnath2019pregroup,abbaszade2021application}. We use the pregroup grammar formalism discussed by \cite{debnath2019pregroup} to define the word interactions and assign the pregroup notations. Further, this work tries to underline a proof of concept using a toy Hindi dataset for binary classification.

\section{Background}

\subsection{Quantum Natural Language Processing}
In Quantum Computing, the information present in qubits is represented in the form of vectors. A similar approach of using vectors (or vector spaces) has also been used frequently in Natural Language Processing involving Word Embeddings such as Word2Vec, and FastText. In recent years, the intersection of Natural Language Processing (NLP) and Quantum Computing has seen notable success, giving rise to a field known as QNLP. This hybrid domain harnesses the potential of Quantum Mechanics to address crucial aspects of language processing, encompassing various NLP tasks. The existing approaches within QNLP range from purely theoretical demonstrations of quantum advantage to practical implementations of algorithms on quantum hardware.
It has many applications, such as machine translation, text summarization, and chatbot creation.
QNLP aims to use Quantum Computers to solve NLP problems. The goal of QNLP is to develop quantum algorithms that can outperform classical algorithms for NLP tasks.
To create an explainable NLP model, we must include the grammatical and syntax of the language in consideration\cite{lambek1958mathematics,chomsky2002syntactic}. To incorporate such properties of language needs the \textit{Distributional Compositional Categorical} (\textit{DisCoCat}) model \cite{coecke2010mathematical,grefenstette2011experimental,sadrzadeh2013frobenius}. Recent works have shown the implementation of QNLP techniques on sentence classification tasks using the DisCoCat model \cite{meichanetzidis2023grammar,metawei2023topic}

\subsection{Pre Group Grammar}
Pregroup grammar, or pregroup formalism, is a mathematical approach for analyzing natural language syntax. It is developed by linguist Joachim Lambek, it offers a more precise and elegant description of linguistic structure than traditional context-free grammars \cite{lambek1958mathematics,lambek1999type} .  In this framework, words are linked to abstract algebraic "types" representing grammatical information, such as nouns and verbs. Rules in pregroup grammar are defined as algebraic operations on these types, enabling structured composition of linguistic expressions. This approach efficiently captures linguistic phenomena, like word order and ambiguity, making it valuable in computational linguistics and linguistic formalism. Pregroup grammar focuses on the algebraic structure within language expressions, attracting research interest in linguistics and computer science. Several languages such as Sanskrit, Arabic, German, French and Japanese have adopted pregroup calculus \cite{cardinal2002algebraic,sadrzadeh2010clitic}:

$$a^{l}\cdot a \rightarrow 1 \rightarrow a \cdot a^{r}$$ 
where $a^{l}$ and $a^{r}$ denote the left and right adjoint of the type of word "a" respectively. For example in the sentence "Ram likes School", subject and object nouns are represented by "n" and the verb "likes" expects a right and left adjoints as these nouns. Algebraic calculation of sentence would lead to flow out of sentence denoted by "s" with  $n^{l}\cdot n \rightarrow 1 \rightarrow n \cdot n^{r}$.

A detailed exploration of the grammatical structures and sentence compositions of English and Hindi reveals a combination of similarities and disparities. Both languages encompass fundamental building blocks such as nouns, verbs, and adjectives, which serve as the foundational elements for constructing sentences. These shared components facilitate basic communication and comprehension across the two languages.
However, where English and Hindi diverge is in the inclusion of specific linguistic features that are unique to each language. For instance, Hindi employs a concept known as "Karak," which plays a crucial role in assigning roles and relationships to different sentence elements. This is a distinct feature absent in English sentence structure. Additionally, tense markers in Hindi such as "Hai", "Thaa" etc convey temporal information differently than their English counterparts, introducing another layer of complexity in the grammatical structure.
To facilitate a better understanding of these differences, Table \ref{tab:pregroup} is presented, illustrating the various pregroup notations used in both languages and their corresponding sentence components. 

\begin{table}[h]
    \centering
     \caption{Pregroup Notations}
    \label{tab:pregroup}
    \begin{tabular}{|c|c|}
    \hline
         \textbf{Symbol} & \textbf{Meaning }   \\
         \hline
         \hline
        $\pi$ & Personal Pronoun \\
        \hline
         \textit{n}& Noun Phrase \\
         \hline
         \textit{p}& Simple Predicate \\
         \hline
         \textit{a}& Adjective \\
         \hline
         \textit{o}& Object or Transitive Verb\\
         \hline
         \textit{$k_1$}& Karaka \\
         \hline
         \textit{$\rho$}& Sambandh \\
         \hline
         \textit{$\alpha$}& Optional Auxiliary \\
         \hline
         \textit{$\tau$}& Tense Marker \\
         \hline
    \end{tabular}
   
\end{table}

\begin{figure*}[ht!]
    
    \centering
    \includegraphics[width= 0.9\textwidth]{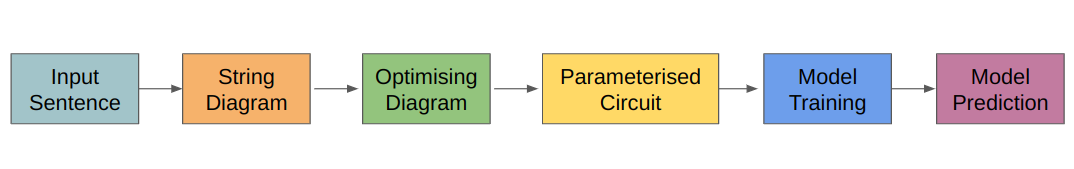}
    \caption{Standard Quantum Natural Language Pipeline}
    \label{fig:pipe}
\end{figure*}

\subsection{Sentence Diagrams}
Sentence diagrams in QNLP are graphical representations that use the principles of Quantum Mechanics to depict the syntactic and semantic structure of sentences in natural language. Unlike traditional sentence diagrams, which typically rely on tree structures to represent grammatical relationships, QNLP sentence diagrams employ quantum-inspired principles to model the complexity and entanglement of linguistic elements within a sentence. In QNLP, sentence diagrams, words, or linguistic units are represented as quantum-like states or vectors in a high-dimensional space, that is like $|\psi\rangle=\alpha|0\rangle+\beta |1\rangle$.  These states can exhibit superposition, entanglement, and other quantum properties, allowing for a more nuanced representation of the relationships between words and their syntactic and semantic roles. The use of quantum principles in sentence diagrams can enable more accurate modeling of phenomena like word ambiguity, context-dependent meaning, and the interaction between words within a sentence. QNLP sentence diagrams are part of the broader effort to apply Quantum Computing and quantum-inspired techniques to natural language understanding, with the goal of enhancing the processing and comprehension of human language.

\subsection{Quantum Circuits From Sentence Diagrams}

To train our NLP model using sentence diagrams and harness the quantum advantage, it's essential to transform these diagrams into trainable parameterised quantum circuits. A specified sequence of gates applied to specific wires in a quantum circuit is known as "\textbf{Ansatz}". To convert a diagram to a circuit, we have several methods or ansatz, including the Instantaneous Quantum Polynomial (IQP) ansatz and Matrix Product States (MPS) approaches. Additionally, there are hybrid Quantum Algorithms introduced earlier \cite{sim2019expressibility}, offering further options for this transformation. These various ansatzes provide us with a range of tools to adapt sentence diagrams into quantum circuits, ultimately enabling us to explore the quantum advantages in natural language processing tasks \cite{tangpanitanon2022explainable,meichanetzidis2020quantum,abbaszadeh2021parametrized,meichanetzidis2023grammar,kartsaklis2021lambeq}.

\section{Methodology}

A standard QNLP Pipeline involves the following steps as described in Figure \ref{fig:pipe}:
\begin{itemize}
    \item \textbf{Input Sentence to String Diagram}: Using pregroup grammar of a language we assign the atomic type to the words and phrases. Then, we use the DisCoCat framework to create string diagrams corresponding to a sentence as shown in Figure 2, wherein the words are connected by cup-shaped wires called "cups" realized using an entangled pair.
    \item \textbf{Optimising String Diagram}: Once the string diagram for a sentence is created, we optimize this diagram by reducing the number of cups so that the dimensionality of the words becomes smaller, here word "jata" needs such correction. This step assists in removing the unnecessary complexities in a sentence diagram.
    \item \textbf{String Diagram to Parametersied Quantum Circuit}: The transition of String Diagrams to Parameterised Quantum Circuit (PQC) carried via standard ansatz such as Instantaneous Quantum Polynomial Ansatz (IQP) or Matrix Products States (MPS) etc.
    \item \textbf{Building and Training Model}: Once the Parameterised Quantum Circuits for each sentence have been built, we can train our prediction model in a similar manner to how we train a classical NLP. Each circuit is assigned initial parameters and then measured. This measurement output is processed further according to the task at hand, such as comparing with the expected output as part of supervised learning. We use a classical optimizer to modify the parameters of the quantum circuits, thus training the model.  
\end{itemize}

\subsection{The Sentences}
We take two simple sentences from the Hindi language to demonstrate how we can build String Diagrams and Parameterised Quantum Circuits from Hindi sentences. These two sentences are:\\
- "Main School Jata hu" which translates to  "I go to school".\\
- "Mukesh ne Khana khaya" which translates to "Mukesh ate the food".
 
\begin{figure}[h!]
    \centering
    \includegraphics[width=0.4\textwidth]{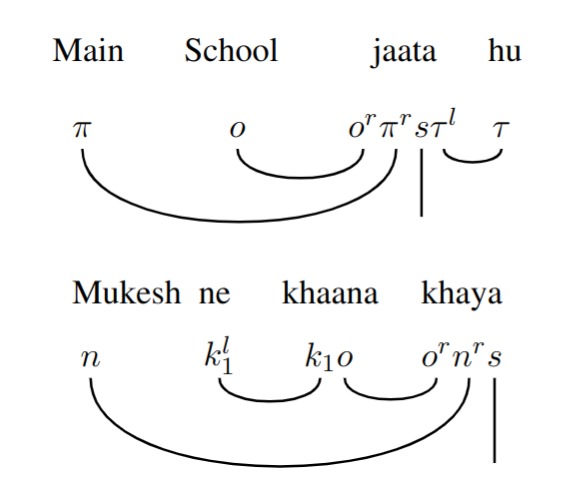}
    \caption{String Diagram for Hindi Sentences}
    \label{fig:string-d}
\end{figure}

\subsection{Creating Sentence Diagram}
\begin{figure}[h!]
    \centering
    \includegraphics[width=0.48\textwidth]{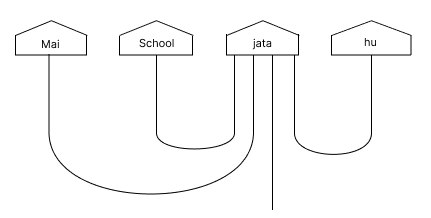}
    \caption{Diagram for Sentence 1, single wired states are of type b and multiple wired are of type d.}
    \label{school}
\end{figure}

\begin{figure}[h!]
    \centering
    \includegraphics[width=0.5\textwidth]{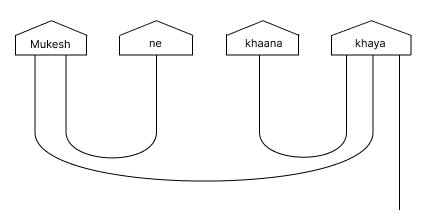}
    \caption{Diagram for Sentence 2, single wired states are of type b and multiple wired are of type d.}
    \label{mukesh}
\end{figure}

Using the pregroup notations and word connection formulated using the pregroup grammar for Hindi, we create the sentence diagrams for the sentences mentioned in Sec 3.1. For the Sentence shown in Fig.\ref{fig:string-d} top, we obtain Figure \ref{school}. Similarly for the Sentence shown in Fig. \ref{fig:string-d} bottom, we obtain the Fig.\ref{mukesh}.

\subsection{Building the Parameterised Quantum Circuit}

Quantum Theory, established as a process theory, employs a diagrammatic language denoted by string diagrams. Specifically, in the experiments we consider, we utilize pure quantum theory. Within this context, processes manifest as unitary operations or quantum gates associated with circuits. The monoidal structure, which facilitates parallel processes, is represented by the tensor product. Sequential composition mirrors the sequence of quantum gates.

\begin{figure}[h!]
    \centering
    \includegraphics[width=0.13\textwidth]{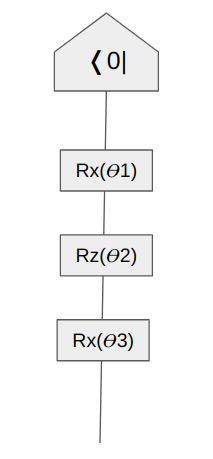}
    \caption{Ansatz for word with single wire}
    \label{fig:single-w}
\end{figure}

\begin{figure}[h!]
    \centering
    \includegraphics[width=0.3\textwidth]{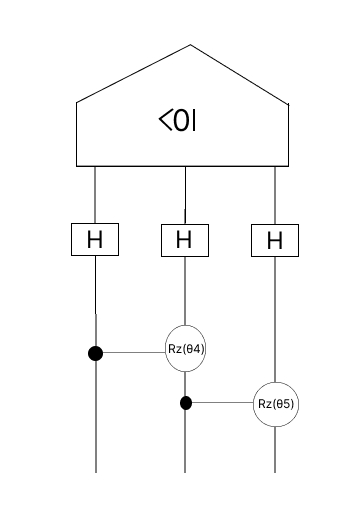}
    \caption{Ansatz for word with multiple wire}
    \label{fig:multi-w}
\end{figure}

Wires assigned the basic pregroup type \( b \) are allocated \( q_b \) qubits. Word-states with a singular output wire obtained from \( \langle 0|\), see Figure \ref{fig:single-w}. To prepare these states, we opt for a series of Unitary gates representing an Euler decomposition. This entails a series of one-qubit unitaries \( R_z(\theta_1) \) to \( R_z(\theta_3)\), which are simple rotational matrices capable of rotating the subjected quantum state vectors. Word-states with multiple output wires evolve from multi-qubit states on \( k \) qubits where \( k > 1 \), which are structured by an IQP-style circuit, see FIgure \ref{fig:multi-w}. The word-circuit is divided into \( d \) layers. Each layer starts with Hadamard gates and then connects every adjacent pair of qubit wires with a  controlled rotational \( CR_z(\theta) \) gate. As all \( CR_z \) gates are commutative, it's feasible to view them as one layer. The Kronecker tensor for \( n \)- single output wires of type \( b \) are connected to word states via a GHZ state. In essence, the GHZ circuit produces the state \( \sum_{a=0}^{2^b} \otimes^n_{i=1} |bin(a) \rangle \), where "bin" is a binary representation. For example $n=2$ correspond to Bell (Entangled) State $|00\rangle+|11\rangle$.The pregroup type \( b \) is mapped to \( q_b \)-many nested Bell effects ($\langle00|+\langle11|$), executed as a CNOT, followed by a Hadamard gate on the control qubit and post-selection on \( |00 \rangle \). This 
\begin{figure}[h!]
    \centering
    \includegraphics[width=0.5\textwidth]{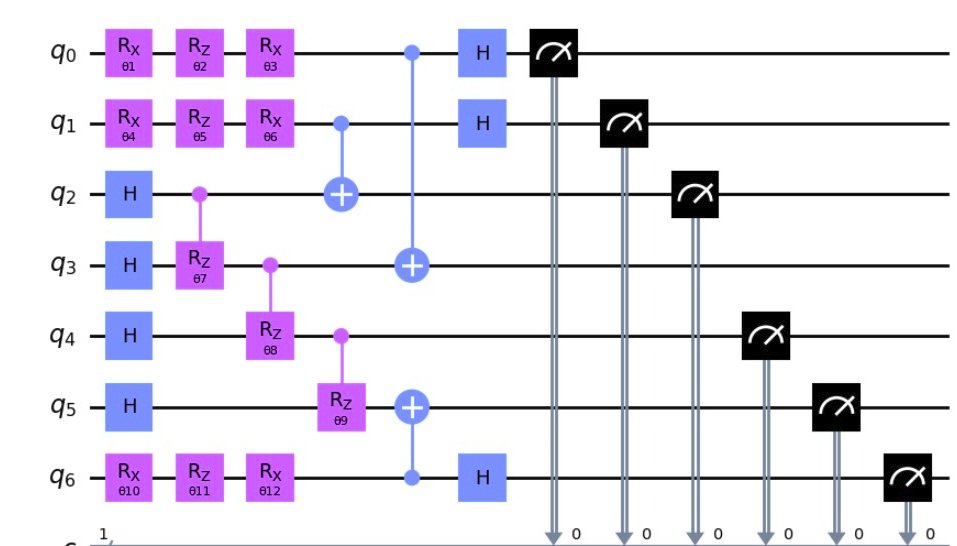}
    \caption{PQC for "Main School Jata hu", end gates are measurements leading to information to classical register "c" shown at the bottom.}
    \label{fig:school}
\end{figure}

\begin{figure}[h!]
    \centering
    \includegraphics[width=0.45\textwidth]{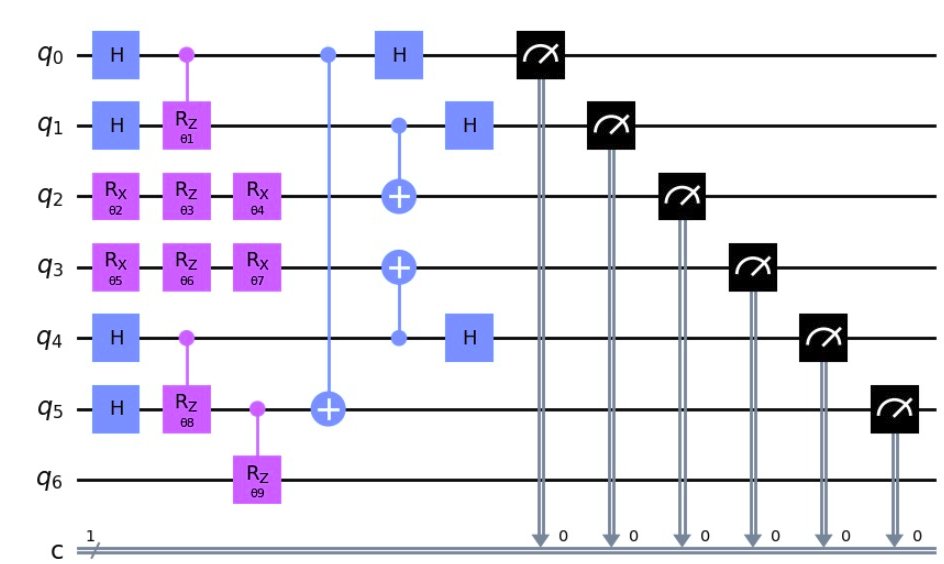}
    \caption{PQC for "Mukesh ne khana khaya", end gates are measurements leading to information to classical register "c" shown at the bottom.}
    \label{fig:mukesh}
\end{figure}

\begin{figure}[h!]
    \centering
    \includegraphics[width =0.5\textwidth]{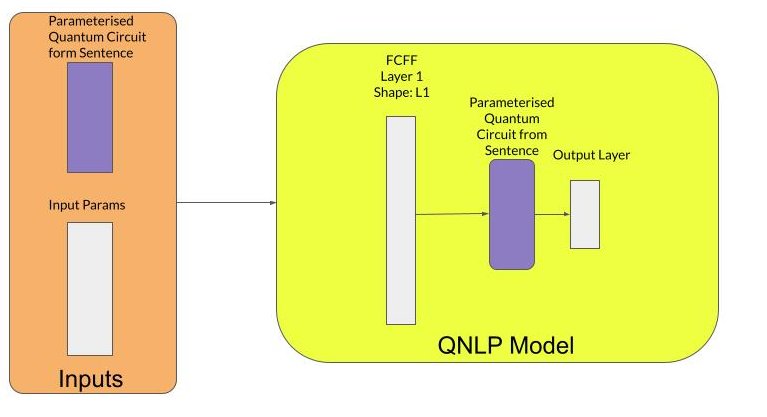}
    \caption{Hybrid Quantum Neural Network for Quantum Natural Language Processing}
    \label{fig:model_hqnn}
\end{figure}

\subsection{Model Training}

The model uses circuits generated from Hindi sentences through the IQP method, along with a vector as input during each forward pass, shown in Figure \ref{fig:model_hqnn}. These sentences result in diverse circuits with varying numbers of parameterised gates, as shown in Figures \ref{fig:mukesh} and \ref{fig:school}. To assign parameters to each rotation gate, we propose a classical-quantum hybrid model. In this setup, the first layer is a Fully Connected Feed Forward (FCFF) network, which expands the input parameters to match the requirements of the parameterised quantum circuit with the highest number of rotation gates among the sentences. Since each circuit demands a different number of input parameters, not all values produced by the Fully Connected Feed Forward layer are utilized for training in each forward pass. The output of this layer is then mapped onto the input values of the parameterised circuit for training. Finally, an output layer with a sigmoid activation function is employed for binary classification.

\begin{figure}[h!]
    \centering
    \includegraphics[width=0.5\textwidth]{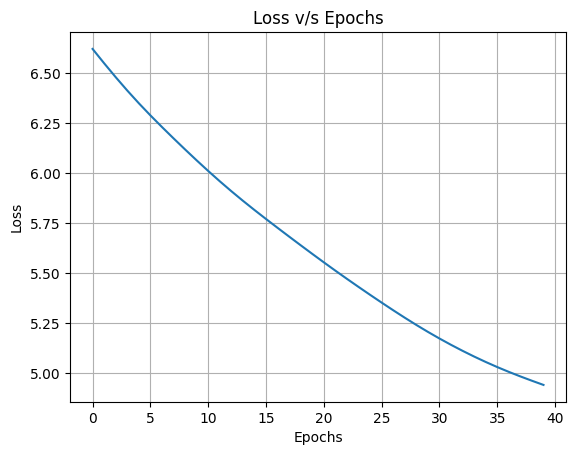}
    \caption{Epoch v/s Loss for training the model on a toy dataset}
    \label{fig:enter-label}
\end{figure}

\section{Discussion and Future work}

In this study, we have delved into the process of constructing trainable parameterized quantum circuits from Hindi sentences, utilizing pre-group grammar and the DisCoCat model. This methodology holds significant relevance within the broader framework of a versatile QNLP Pipeline. Once we successfully create parameterized quantum circuits for a curated dataset comprising Hindi sentences, the next crucial step involves training these circuits to develop a proficient QNLP Model. Further, we proposed a hybrid classical-quantum approach to train a Quantum Natural Language Model, implementing it on a toy dataset.  
Looking forward, our research aspirations encompass an extension of this work to devise a nuanced sentence classifier for Hindi sentences, integrating topic and context awareness through a QNLP approach. We draw inspiration from a study by \cite{metawei2023topic}, aiming to align our approach with their advancements in this domain. However, we acknowledge the inherent complexities lying ahead in this endeavour. Hindi exhibits a notably freer word order compared to English, presenting a challenge in accommodating potential restrictions on word order movements within the construction of the PQC. Addressing and overcoming these intricacies are essential components of our ongoing and future research, paving the way for a more refined and accurate QNLP approach tailored to the nuances of the Hindi language.

\section{Conclusion}
In conclusion, this paper presents a significant advancement in the field of QNLP by extending its applications to the Hindi language. By leveraging the pregroup representation of Hindi and the DisCoCat framework, we have successfully designed sentence diagrams, which are then translated into Parameterised Quantum Circuits using the IQP style ansatz. The resulting parameterised circuits enable the development of grammar-aware and topic-aware sentence classifiers tailored to the nuances of the Hindi language. The hybrid classical-quantum model enables the training over hindi sentences with the help quantum natural language processing. This novel approach not only contributes to the growing body of QNLP research but also addresses the need for more explainable NLP systems in languages beyond English, which promises exciting possibilities for natural language understanding and processing across diverse linguistic landscapes.

\end{document}